%% file: main.tex
\documentclass[10pt,twocolumn,letterpaper]{article}
\usepackage[letterpaper,margin=1in,columnsep=0.3125in]{geometry}
\input{preamble}
\definecolor{linkblue}{rgb}{0.21,0.49,0.74}
\usepackage[breaklinks,colorlinks,allcolors=linkblue]{hyperref}
\title{AgentMoT: An Agentic Mixture-of-Transformers\\for End-to-End Autonomous Driving}
\input{authors}
\date{}
\hypersetup{pdftitle={AgentMoT: An Agentic Mixture-of-Transformers for End-to-End Autonomous Driving},pdfauthor={\arxivauthornames}}
\begin{document}
\twocolumn[{\preprinttitle\input{fig/0_teaser}}]
\arxivauthornotes
\input{sec/0_abstract}
\input{sec/1_introduction}
\input{sec/2_related_work}
\input{sec/3_awa_drive}
\input{sec/4_dual_loop_evolution}
\input{sec/5_experiments}

\input{sec/6_conclusion}
\label{maintextend}
\input{sec/llm_usage}
{\small
\bibliographystyle{ieeenat_fullname}
\bibliography{main}
}
\clearpage
\appendix
\setcounter{table}{0}
\setcounter{figure}{0}
\setcounter{equation}{0}
\renewcommand{\thetable}{S\arabic{table}}
\renewcommand{\thefigure}{S\arabic{figure}}
\renewcommand{\theequation}{S\arabic{equation}}
\renewcommand{\theHtable}{appendix.\arabic{table}}
\renewcommand{\theHfigure}{appendix.\arabic{figure}}
\renewcommand{\theHequation}{appendix.\arabic{equation}}
\raggedbottom
\twocolumn[{\begin{center}\Large\bfseries Appendix\end{center}\vspace{0.5em}}]
\noindent This appendix specifies the model configuration, collection records, offline learning procedures, and evaluation protocol for \awa{}.
\input{supp/1_model_and_data}
\input{supp/2_router}

\input{supp/3_agent}
\input{supp/4_evaluation}
\end{document}

%% file: preamble.tex
\usepackage[T1]{fontenc}
\usepackage{times}
\usepackage[numbers,sort&compress]{natbib}
\usepackage[hyphens]{url}
\usepackage[font=small,labelsep=period]{caption}
\usepackage{enumitem}
\usepackage{graphicx}
\usepackage{amsmath,amssymb}
\usepackage{booktabs,multirow,tabularx}
\usepackage{xcolor}
\usepackage{microtype}
\newcommand{\awa}{AgentMoT}
\newcommand{\native}{\textsc{Native}}

\DeclareMathOperator{\KL}{KL}
\newcommand{\E}{\mathbb{E}}

\setlist[itemize]{nosep,leftmargin=*}

\makeatletter
\newcommand{\preprinttitle}{%
  \begin{center}
    \vspace*{13pt}
    {\Large\bfseries \@title\par}
    \vspace{12pt}
    {\large\begin{tabular}{c}\@author\end{tabular}\par}
    \vspace{12pt}
  \end{center}}
\renewcommand{\section}{\@startsection{section}{1}{0pt}%
  {-10pt plus -2pt minus -2pt}{7pt}{\normalfont\large\bfseries}}
\renewcommand{\subsection}{\@startsection{subsection}{2}{0pt}%
  {-8pt plus -2pt minus -2pt}{5pt}{\normalfont\fontsize{11}{13}\selectfont\bfseries}}
\renewcommand{\@seccntformat}[1]{\csname the#1\endcsname.\enspace}
\makeatother

\renewenvironment{abstract}
  {\begin{center}\large\bfseries\abstractname\end{center}\noindent\itshape\ignorespaces}
  {\par\medskip}

%% file: authors.tex
\newcommand{\arxivauthornames}{Qihang Huang, Collister Chua, YuJie Yan, Shiyu Fang, Zhan Chen, Haoran Xu, Zirui Li, Chen Lv}
\author{%
Qihang Huang\textsuperscript{1*}\quad
Collister Chua\textsuperscript{1*}\quad
YuJie Yan\textsuperscript{1}\quad
Shiyu Fang\textsuperscript{2}\\
Zhan Chen\textsuperscript{1}\quad
Haoran Xu\textsuperscript{1}\quad
Zirui Li\textsuperscript{1}\quad
Chen Lv\textsuperscript{1\textdagger}\\[0.3em]
\textsuperscript{1}Nanyang Technological University\quad
\textsuperscript{2}Tongji University}

\newcommand{\arxivauthornotes}{%
  \begingroup
  \renewcommand{\thefootnote}{\fnsymbol{footnote}}%
  \footnotetext[1]{Equal contribution.}%
  \footnotetext[2]{Corresponding author: \href{mailto:lyuchen@ntu.edu.sg}{lyuchen@ntu.edu.sg}.}%
  \endgroup}

%% file: fig/0_teaser.tex
\begin{center}
\includegraphics[width=\textwidth]{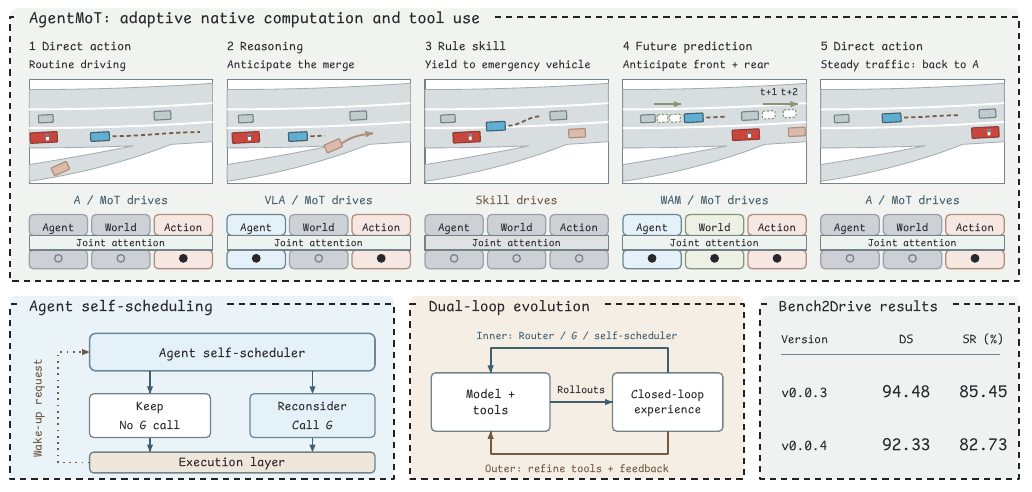}
\captionsetup{type=figure,hypcap=false}
\caption{\textbf{Adaptive computation and tool use during a drive.}
VLA reasons about a ramp vehicle merging ahead. $G$ then selects the rule skill to yield to an emergency vehicle. WAM predicts surrounding traffic (hollow vehicles), and A resumes steady driving. The blue vehicle is ego, the red vehicle is the emergency vehicle, and dashed paths are ego plans. Dots indicate active native experts, all grey while the rule skill controls the vehicle. Below, the self-scheduler screens ordinary wake-up requests. Mandatory requests call $G$ directly. Dual-loop evolution updates scheduling, skills, and execution feedback.}
\label{fig:teaser}
\end{center}

%% file: sec/0_abstract.tex
\begin{abstract}
Driving models need to choose which capabilities to use and when to reconsider their decisions. We introduce \awa{}, a Mixture-of-Transformers with built-in agent reasoning and system scheduling. Combinations of agent, world, and action experts form three jointly trained pathways for direct-action, vision-language-action, and world-action computation. The Router selects a native pathway, while the agent expert supplies reasoning and selects skills. Execution feedback raises wake-up requests, and the agent self-scheduler learns whether to reconsider the task decision or keep it. Mandatory requests invoke the agent directly. The relative-capability principle guides these choices by relating the benefit of additional computation or intervention to its cost. After open-loop initialization, dual-loop evolution trains the Router, agent, and self-scheduler from closed-loop experience while failure-driven updates refine skills and execution feedback. \awa{} achieves state-of-the-art results on Bench2Drive v0.0.3 and v0.0.4, with 94.48/92.33 driving score and 85.45/82.73\% success rate, respectively. With skills disabled, learned routing improves over the strongest fixed pathway by 1.79 driving-score points and 4.09 percentage points of success rate on v0.0.3. Ablation studies further show gains from closed-loop agent learning, two-view queries, and event-triggered wake-up.
\end{abstract}

%% file: sec/1_introduction.tex
\section{Introduction}
\label{sec:intro}
End-to-end driving connects sensor observations to motion through learned perception and planning~\cite{uniad,vad}. Vision-language reasoning enriches scene interpretation~\cite{simlingo,automot}, while world prediction anticipates traffic evolution~\cite{occwam,interact}. Driving agents further provide access to specialized tools~\cite{drivingagent,driveagentr1,autoagent0}. Together, these capabilities expand what a driving model can do. Their value in a particular scene depends on which capabilities the model chooses to use and when it revisits that choice.

Using these capabilities requires choosing a native pathway, deciding whether to delegate to a skill, and determining when to reconsider (Fig.~\ref{fig:teaser}). Each choice depends on context. Reasoning and prediction cost computation. A skill suited to a blocked lane may be inappropriate for an ordinary stopped vehicle. Repeated confirmation wastes agent calls, while new interactions demand attention.

\input{fig/1_overview_v6}

We present \textbf{\awa{}}, which embeds agent capabilities into a driving Mixture-of-Transformers (MoT, Fig.~\ref{fig:overview}). A Router selects among direct-action (A), vision-language-action (VLA), and world-action model (WAM) pathways formed by agent, world, and action experts. The agent expert supplies reasoning within VLA and WAM. System scheduling combines task selection by $G$ with call timing from the agent self-scheduler. $G$ selects native driving or a skill and target. The execution layer builds the skill--target menu and sends wake-up requests with updated driving context. The agent self-scheduler screens ordinary requests, choosing reconsider or keep. Mandatory requests call $G$ directly and can add a rear view to identify an approaching vehicle before deciding whether to yield.

Under the \emph{relative capability} principle, an option is useful when its expected benefit over the default justifies its cost. The appropriate choice depends on both the scene and the capabilities available to the model. This motivates updating scheduling as those capabilities evolve.

\textbf{Dual-loop evolution} keeps these learned choices aligned with the available capabilities. After joint native training and open-loop initialization of the Router and $G$, the inner loop trains the Router, $G$, and the agent self-scheduler on closed-loop training-route experience. The outer loop attributes failures and refines skills and execution feedback. Rollouts with the updated model and skills supply experience for the next round.

Our contributions are threefold. \textbf{(1)} We introduce \awa{}, which unifies three native driving paradigms with a shared agent expert for reasoning and tool use. The model learns to select native computation, delegate tasks, and reconsider decisions in response to execution feedback. \textbf{(2)} We develop dual-loop evolution that connects scheduling learned from closed-loop experience with failure-driven refinement of skills and execution feedback. \textbf{(3)} We achieve state-of-the-art performance on Bench2Drive v0.0.3 and v0.0.4. Ablation studies demonstrate the benefits of learned routing, agent learning, and execution feedback.

%% file: fig/1_overview_v6.tex
\begin{figure*}[t]
\centering
\includegraphics[width=\textwidth]{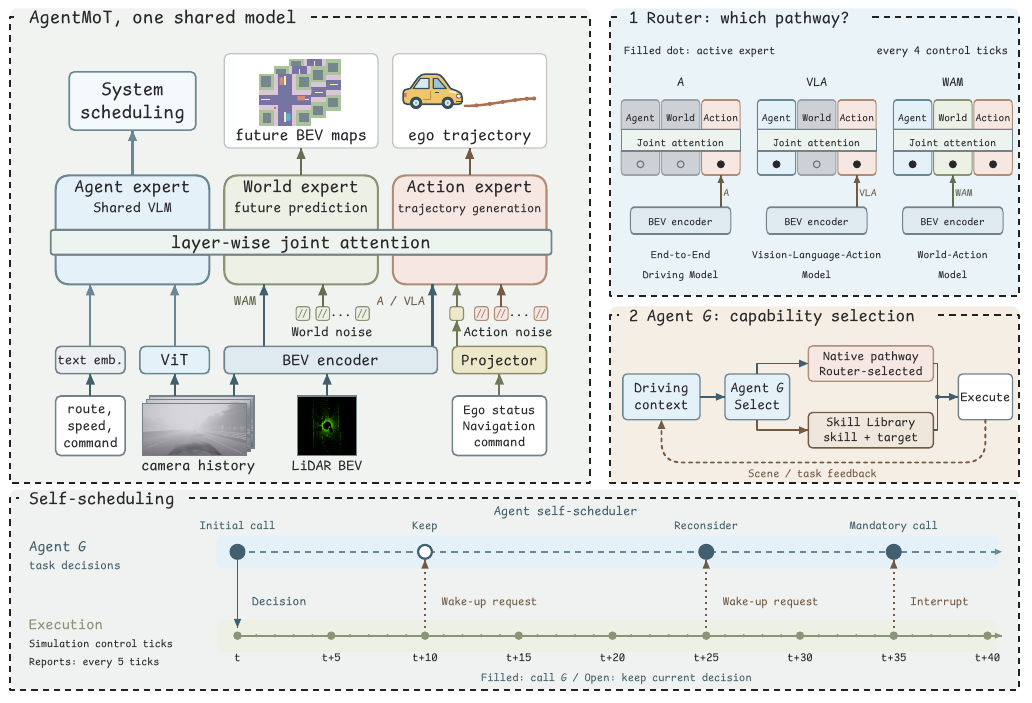}
\caption{\textbf{\awa{} selects computation, tools, and when to call its agent.}
Left, system scheduling covers task selection and self-scheduling. Attention passes agent context to world and action experts, and world context to action. Native reasoning and $G$ use role-specific LoRA adapters. Top right, the Router independently selects A, VLA, or WAM from BEV tokens and context before native expert activation. Middle right, $G$ selects native driving or a skill and target. Bottom, the self-scheduler screens ordinary wake-up requests. Mandatory requests call $G$ directly. Decisions flow down and execution feedback flows up. Larger execution markers denote state reports. Time is in simulation control ticks, and sensor images are recorded observations.}
\label{fig:overview}
\end{figure*}

%% file: sec/2_related_work.tex
\section{Related Work}
\label{sec:related}
\paragraph{End-to-end driving.}
End-to-end driving integrates perception, prediction, and planning through learned scene representations~\cite{uniad,vad,drivetransformer}. Vision-language models enrich action generation with semantic understanding and reasoning~\cite{simlingo,autovla}. Future-scene prediction improves planning~\cite{wote,unidrivewm}, with INTERACT~\cite{interact} learning interaction dynamics from driving videos. Closed-loop post-training further improves driving policies~\cite{craft,minddrive,takevla}. Together, these directions motivate the complementary direct-action, reasoning, and prediction capabilities in \awa{}.

\paragraph{Mixture-of-Transformers.}
MoT combines modality-specific Transformer parameters with shared attention, allowing specialized representations to interact within one model~\cite{liang2025mot}. AutoMoT~\cite{automot} couples understanding and action experts through layer-wise attention for driving. OccWAM~\cite{occwam} uses a three-expert MoT to jointly generate future occupancy and ego trajectories. These architectures share context across specialized forms of computation. \awa{} organizes agent, world, and action experts into three selectable driving pathways. The agent expert also supports task decisions through the shared vision-language backbone, connecting native planning with tool use and system scheduling.

\paragraph{Adaptive computation and agents.}
Adaptive computation allocates resources according to the decision, from selecting specialized experts~\cite{drivemoe} to adjusting reasoning effort~\cite{adathinkdrive,thinkless}. Temporal decision methods additionally learn when to observe or act~\cite{biedenkapp2021temporl,zhou2024timing}. Tool-using agents connect reasoning to environmental feedback~\cite{react}. Driving agents extend this approach with module scheduling, active perception, and verification during execution~\cite{drivingagent,driveagentr1,autoagent0}. These ideas address both capability choice and the timing of decisions. In \awa{}, the Router selects native computation, $G$ chooses a task, and the self-scheduler determines whether execution feedback warrants reconsideration.

%% file: sec/3_awa_drive.tex
\section{\awa{}}
\label{sec:model}
\awa{} learns which native pathway to use, whether to delegate a task, and when to reconsider the task decision. The model generates native plans and task decisions end to end, with skills invoked as tools. Figure~\ref{fig:overview} connects these choices to execution feedback.

\subsection{Relative capability}
\label{sec:relative}
For context $x$, option $a$, and default $a_0$, define
\begin{equation}
 \Delta(x,a;a_0)=[B(x,a)-B(x,a_0)]-[C(x,a)-C(x,a_0)],
 \label{eq:relative}
\end{equation}
where $B$ is expected driving benefit and $C$ is computation or intervention cost over the decision's horizon, expressed in compatible units. Positive relative capability motivates using the additional capability. The defaults are A for routing, native driving for delegation, and keeping the current task decision for self-scheduling.

This principle guides three learning problems. Routing explicitly compares plan rewards and latency costs. Delegation uses task applicability as supervision, while self-scheduling balances call reduction against retaining calls labeled for reconsideration. Section~\ref{sec:evolution} describes their training signals and objectives.

\subsection{Three native pathways}
A bird's-eye-view (BEV) encoder fuses RGB and LiDAR into spatial features. Camera history, speed, and navigation inputs enter the agent expert through its vision-language backbone. The pathways share the action expert's Transformer layers and use pathway-specific input and output modules. They differ in which expert representations condition trajectory generation.

\textbf{A (direct action).} The action expert maps BEV features, ego state, and navigation context to an ego trajectory. \textbf{VLA (vision-language-action).} The agent expert's vision-language representations supply layer-wise key--value context to action generation. \textbf{WAM (world-action model).} Agent reasoning conditions the world expert, which predicts six future BEV semantic maps at 0.5-second intervals. The action expert attends to both agent and world key--value states. Obstacle classes in the predicted maps and perceived object boxes guide trajectory refinement for clearance and smooth motion. A uses the action expert, VLA adds the agent expert, and WAM uses all three.

Each pathway generates six trajectory points over three seconds. A lightweight Router reads BEV tokens and compact context features (ego motion, navigation, and situation indicators) before activating the selected experts. The shared trajectory interface lets the model switch computation as the scene changes.

\subsection{Agent reasoning and system scheduling}
\label{sec:agent}
System scheduling combines task selection by $G$ with call timing from the agent self-scheduler (Sec.~\ref{sec:timing}). The agent expert supplies native reasoning and generates task decisions through $G$. Calls to $G$ add structured scene and task information and a skill--target menu to the image inputs. Native reasoning and $G$ share the vision-language backbone with role-specific low-rank adaptation (LoRA) adapters and separate key--value caches. Agent post-training updates the task adapter while keeping the backbone and native-planning parameters fixed.

The \emph{skill library} provides reusable task-level tools for traffic negotiation, obstruction passing, navigation-goal adaptation, and traffic-rule compliance. The rule skill includes emergency-vehicle yielding. The execution layer constructs a \emph{skill--target menu} from perceived objects, navigation context, and task state. At time $t$, the task-selection policy is $\pi_G(\cdot\mid x_t^G,\mathcal M_t)$, where the \emph{driving context} $x_t^G$ contains images, perception-derived scene information, and task state, and $\mathcal M_t$ is the skill--target menu. $G$ chooses $\native$ (native driving) or a skill and target. For $\native$, the Router-selected pathway plans the trajectory. The execution layer checks skill--target choices, applies accepted choices through controls or planning references, and reports task progress to later calls.

\subsection{Self-scheduling from execution feedback}
\label{sec:timing}
Task decisions often remain useful across many planning steps. The \emph{execution layer} refreshes the driving context and skill--target menu while native planning or a skill controls the vehicle. It sends \emph{wake-up requests} when observations warrant a new decision. Ordinary requests come from menu changes, task and traffic events, and periodic reviews. The \emph{agent self-scheduler} screens these requests, choosing \emph{reconsider} (call $G$) or \emph{keep} (reuse the current task decision).

Mandatory requests bypass this choice and call $G$ directly with refreshed inputs. Execution interrupts and conflicts between the current task and observed conditions raise mandatory requests. Active wake-up rules also raise mandatory requests for persistent lack of progress, a new persistent blocker, or persistent new skill options. An age-triggered refresh makes an existing request mandatory once a preset decision-age threshold is reached. The execution layer supplies the evidence and requests reconsideration, while $G$ makes the task decision.

\paragraph{Two-view queries.}
\label{sec:two_view}
Execution feedback can also request additional visual evidence. A rear-vehicle cue can trigger a mandatory \emph{two-view query} with current front and rear images. The answer confirms the vehicle type and whether yielding is needed.

%% file: sec/4_dual_loop_evolution.tex
\section{Dual-Loop Evolution}
\label{sec:evolution}
The value of a routing or task decision depends on the capabilities it selects. The inner loop trains the Router, $G$, and agent self-scheduler from paired plan rewards, task applicability, and agent-call outcomes. The outer loop uses failures to refine skills and execution feedback. Rollouts with the updated model and skills provide the experience for the next round (Fig.~\ref{fig:evolution}).

\input{fig/2_evolution}

\subsection{From joint training to closed-loop evolution}
\label{sec:rule}
Open-loop training mixes A, VLA, and WAM driving examples with SimLingo question--answer supervision~\cite{simlingo} and GPT-annotated skill--target choices for $G$. This establishes the native pathways and connects scene understanding to skill selection. The open-loop Router learns from pathway trajectory errors against PDM-Lite reference plans.

Training-route rollouts then produce \emph{closed-loop records}. They supply paired pathway rewards, recorded $G$ proposals, and agent-call outcomes to the inner-loop objectives below, and failure episodes to the outer loop.

\input{tab/1_bench2drive}
\input{tab/12_bench2drive_v004}
\input{tab/2_abilities}

\subsection{Learning which pathway to use}
\label{sec:router_learning}
Router learning uses the one-step decision structure of \emph{contextual bandits} in reinforcement learning (RL)~\cite{levine2020offline,brandfonbrener2021offline,dudik2011doubly}. The Router optimizes the reward of the plan selected at each recorded context. For each recorded observation, we generate three-second plans from A, VLA, and WAM with paired seeds. Scoring all three provides full-information supervision for offline learning with reward
\begin{equation}
 r^R(x,k)=\overline S_{\rm plan}(x,k)
 -\lambda_d\min(\overline d(x,k),3)-\lambda_c\ell_k.
 \label{eq:router_reward}
\end{equation}
Here $r^R$ is the Router reward for context $x$ and pathway $k\in\{\mathrm A,\mathrm{VLA},\mathrm{WAM}\}$. $\overline S_{\rm plan}$ is the change in route driving score over a three-second plan replay, computed with official driving criteria and recorded scene evolution. $\overline d$ is expert-trajectory deviation in meters, and $\ell_k$ is latency in seconds. Overbars average four paired seeds. The weights $\lambda_d$ and $\lambda_c$ penalize deviation (capped at 3\,m) and latency.

The Router makes two binary choices: whether to leave the default A for deeper computation, and then VLA or WAM. Both are trained on Eq.~\ref{eq:router_reward}. The first minimizes expected regret, the reward lost relative to the better of A and deeper computation under the current second choice. This objective favors deeper computation when its relative capability is positive (Eq.~\ref{eq:relative}). The second predicts the higher-reward pathway between VLA and WAM, weighted by their reward gap on contexts where deeper computation improves on A. A refinement stage then revisits contexts assigned to A and moves them to VLA when a comfort-sensitive reward favors it. Appendix~\ref{supp:router} details the paired training targets and optimization settings.

\subsection{Learning whether to delegate}
\label{sec:g_learning}
Routing exploits complementary native capabilities, but their failure histories overlap. In native-pathway runs on 139 training routes, 61 routes contain at least one failed run (DS below 100). On 32 of them (52.5\%), A, VLA, and WAM have each failed at least once. These shared difficulties motivate skills with explicit geometry and motion models for constrained optimization, such as traffic negotiation and obstruction passing. With native driving as the default, the agent must learn whether a skill applies and which target it should act on.

\paragraph{Agent post-training.}
We adapt $G$ with closed-loop task supervision and reference regularization that retains its prior responses. Offline labeling rules use scenario annotations, the recorded skill--target menu, and task state to construct answer targets. These targets retain applicable recorded answers or remove inapplicable skill calls. Two-view examples supply vehicle-identification and yielding targets and use only the answer loss.

We mix closed-loop records with replayed reference driving examples and optimize
\begin{equation}
\begin{split}
 \mathcal L_G={}&\mathcal L_{\rm ans}+\alpha\mathcal L_{\rm first}\\
 &+\beta_{\rm cl}\mathcal L_{\rm KL}^{\rm cl}
 +\beta_{\rm replay}\mathcal L_{\rm KL}^{\rm replay}.
\end{split}
\label{eq:g_posttraining}
\end{equation}
Here $\mathcal L_{\rm ans}$ is full-answer cross-entropy. The first-token term $\mathcal L_{\rm first}$ uses KL divergence to match a soft target that downweights inapplicable skills. The two reference KL terms use the fixed pre-update agent to regularize responses on closed-loop records and replayed driving examples, respectively. The weights $\alpha$, $\beta_{\rm cl}$, and $\beta_{\rm replay}$ balance these terms. Appendix~\ref{supp:agent} specifies target construction and loss weights.

\subsection{Learning when to reconsider}
\label{sec:timing_learning}
Execution raises wake-up requests on events and periodic reviews. Most resulting agent calls retain the current task decision (Table~\ref{tab:consultations}). With keep as the default, the agent self-scheduler learns to reserve calls for attempted revisions. It uses feedback available before the call to predict whether reconsideration is needed. Calls logged before learned self-scheduling provide supervision. A call receives the reconsider label ($m_j=1$) if it changes accepted skill--target choices or returns only rejected proposals. All other calls receive the keep label ($m_j=0$). This retains attempted revisions even when the execution layer rejects them. Appendix~\ref{supp:timing} defines the outcome categories.

The pre-call feedback $x_j$ contains the wake-up request, elapsed time, scene dynamics, task state, and previous answer. The self-scheduler is a lightweight classifier with parameters $\phi$ that predicts a reconsider probability $p_\phi(x_j)$. We train it with weighted binary cross-entropy,
\begin{equation}
 \mathcal L_{\rm sched}=-\E_j[
 w_+m_j\log p_\phi(x_j)+(1-m_j)\log(1-p_\phi(x_j))].
 \label{eq:timing_objective}
\end{equation}
The weight $w_+$ balances positive and negative examples. An ordinary wake-up request calls $G$ when $p_\phi(x_j)\geq\tau$, where $\tau$ is a calibrated threshold. Otherwise, the current task decision is kept. Mandatory requests call $G$ directly (Sec.~\ref{sec:timing}).

\subsection{Outer-loop attribution and capability repair}
\label{sec:outer}
GPT reviews failed training episodes with trajectories, scheduling choices, task states, and official infraction events. It distinguishes errors in pathway selection and agent decisions from errors in skill execution. Skill-execution errors guide coding-agent changes to skills or the execution layer, including the feedback and wake-up path. The authors review these changes before training-route rollouts generate records for the next inner-loop update.

%% file: fig/2_evolution.tex
\begin{figure*}[t]
\centering
\includegraphics[width=\textwidth]{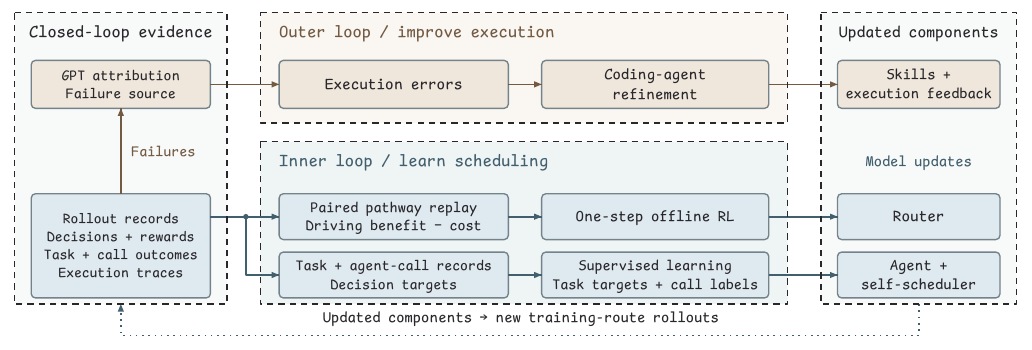}
\caption{\textbf{Dual-loop evolution.} Closed-loop records supply failure evidence and learning targets. The outer loop uses GPT attribution and coding-agent refinement to update skills and execution feedback. The inner loop updates the Router from paired plan rewards, agent task decisions from task applicability, and the self-scheduler from agent-call outcomes. The dotted path collects new training-route rollouts with the updated components.}
\label{fig:evolution}
\end{figure*}

%% file: tab/1_bench2drive.tex
\begin{table*}[!t]
\centering
\caption{\textbf{Closed-loop driving performance on Bench2Drive v0.0.3.} Expert denotes the policy used to collect training data. Baselines follow the original papers. Best and second-best reported scores are bold and underlined, respectively.}
\label{tab:main}
\small
\begin{tabular*}{\textwidth}{@{\extracolsep{\fill}}llcrr@{}}
\toprule
Method & Venue & Expert & DS $\uparrow$ & SR (\%) $\uparrow$ \\
\midrule
WoTE~\cite{wote} & ICCV'25 & Think2Drive & 61.71 & 31.36 \\
AutoVLA~\cite{autovla} & NeurIPS'25 & PDM-Lite & 78.84 & 57.73 \\
UniDrive-WM~\cite{unidrivewm} & ECCV'26 & Think2Drive & 79.31 & 56.42 \\
MindDrive-L~\cite{minddrive} & ECCV'26 & Think2Drive & 80.59 & 58.26 \\
SimLingo~\cite{simlingo} & CVPR'25 & PDM-Lite & 85.07 & 67.27 \\
DeepSight~\cite{deepsight} & ICML'26 & Think2Drive & 86.23 & 71.36 \\
BridgeDrive~\cite{bridgedrive} & ICLR'26 & PDM-Lite & 87.99 & 74.99 \\
KnowVal-SimLingo~\cite{knowval} & CVPR'26 & PDM-Lite & 88.42 & 69.03 \\
AutoMoT+~\cite{automot} & ICML'26 & PDM-Lite & 89.42 & 74.09 \\
TakeVLA~\cite{takevla} & arXiv'26 & PDM-Lite & 89.72 & 73.73 \\
OccWAM~\cite{occwam} & arXiv'26 & PDM-Lite & \underline{90.45} & \underline{77.27} \\
\midrule
\textbf{\awa{}} & --- & PDM-Lite & \textbf{94.48} & \textbf{85.45} \\
\bottomrule
\end{tabular*}
\end{table*}

%% file: tab/12_bench2drive_v004.tex
\begin{table}[!t]
\centering
\caption{\textbf{Closed-loop driving performance on Bench2Drive v0.0.4.} Baselines follow the official release~\cite{bench2drive004}. Best and second-best reported scores are bold and underlined, respectively.}
\label{tab:main_v004}
\footnotesize
\begin{tabular*}{\linewidth}{@{\extracolsep{\fill}}lrr@{}}
\toprule
Method & DS $\uparrow$ & SR (\%) $\uparrow$ \\
\midrule
ADMLP~\cite{admlp} & 35.39 & 0.00 \\
VAD~\cite{vad} & 38.65 & 17.27 \\
UniAD~\cite{uniad} & 38.69 & 19.09 \\
TCP~\cite{tcp} & 51.44 & 26.36 \\
ThinkTwice~\cite{thinktwice} & 52.84 & 27.73 \\
DriveAdapter~\cite{driveadapter} & 54.96 & 30.45 \\
DriveTransformer~\cite{drivetransformer} & 60.20 & 35.91 \\
Drive-$\pi_0$~\cite{drivemoe} & 69.71 & 45.91 \\
SimLingo~\cite{simlingo} & \underline{86.55} & \underline{70.45} \\
\midrule
\textbf{\awa{}} & \textbf{92.33} & \textbf{82.73} \\
\bottomrule
\end{tabular*}
\end{table}

%% file: tab/2_abilities.tex
\begin{table}[!t]
\centering
\caption{\textbf{Multi-ability success rates (\%).} Mean averages the five official abilities. Baselines follow published results~\cite{bridgedrive} for v0.0.3 and the official release~\cite{bench2drive004} for v0.0.4.}
\label{tab:abilities}
\footnotesize
\setlength{\tabcolsep}{1.5pt}
\centering
\textbf{Bench2Drive v0.0.3}\par\smallskip
\begin{tabular*}{\linewidth}{@{\extracolsep{\fill}}lrrrrrr@{}}
\toprule
Method & Over. & Merge & Brake & \shortstack{Give\\way} & Signs & Mean \\
\midrule
DriveTransformer~\cite{drivetransformer} & 35.00 & 17.57 & 48.36 & 40.00 & 52.10 & 38.60 \\
DriveMoE~\cite{drivemoe} & 40.00 & 34.67 & 65.45 & 40.00 & 59.44 & 47.91 \\
MindDrive-L~\cite{minddrive} & 66.67 & 42.31 & 76.67 & 50.00 & 68.42 & 60.81 \\
TransFuser++~\cite{hiddenbiases} & 57.77 & 58.75 & 83.33 & 40.00 & 82.11 & 64.39 \\
SimLingo~\cite{simlingo} & 57.04 & 54.01 & 88.33 & 53.33 & 82.45 & 67.03 \\
TakeVLA~\cite{takevla} & 64.44 & 63.64 & 91.67 & 50.00 & 85.48 & 71.05 \\
BridgeDrive~\cite{bridgedrive} & 66.67 & 69.92 & 90.00 & 50.00 & 89.47 & 73.15 \\
\midrule
\textbf{\awa{}} & \textbf{88.89} & \textbf{78.75} & \textbf{95.00} & \textbf{90.00} & \textbf{94.74} & \textbf{89.48} \\
\bottomrule
\end{tabular*}
\par\medskip
\centering
\textbf{Bench2Drive v0.0.4}\par\smallskip
\begin{tabular*}{\linewidth}{@{\extracolsep{\fill}}lrrrrrr@{}}
\toprule
Method & Over. & Merge & Brake & \shortstack{Give\\way} & Signs & Mean \\
\midrule
DriveTransformer & 30.43 & 39.06 & 49.12 & 0.00 & 61.63 & 36.05 \\
ThinkTwice & 20.29 & 20.31 & 59.64 & 33.33 & 66.86 & 40.09 \\
DriveAdapter & 26.09 & 21.88 & 64.90 & 33.33 & 69.18 & 43.08 \\
Drive-$\pi_0$ & 24.64 & 39.06 & 73.68 & 33.33 & 66.86 & 47.52 \\
SimLingo & 63.77 & 60.94 & 91.23 & 50.00 & 90.70 & 71.34 \\
\midrule
\textbf{\awa{}} & \textbf{81.16} & \textbf{73.44} & \textbf{92.98} & \textbf{83.33} & \textbf{93.02} & \textbf{84.79} \\
\bottomrule
\end{tabular*}
\end{table}

%% file: sec/5_experiments.tex
\section{Experiments}
\label{sec:experiments}

\input{tab/10_paradigms}
\input{tab/3_ablations}
\input{fig/3_paradigms}
\input{tab/11_consultations}
\input{fig/4_police_case}

\subsection{Experimental setup}
\label{sec:implementation}
\paragraph{Benchmark and data.}
We evaluate in CARLA 0.9.15~\cite{carla} on Bench2Drive v0.0.3 and v0.0.4~\cite{bench2drive,bench2drive004}. Each version contains 220 evaluation routes. Version 0.0.4 introduces a new evaluation set with different routes and scenario instances~\cite{bench2drive004}. We report driving score (DS), success rate (SR, the percentage of routes with DS equal to 100), and the five official ability groups separately for each version. We initialize the model from driving logs and refine routing, task selection, and self-scheduling with closed-loop CARLA records. Following the benchmark protocol, training routes are separate from the 220 evaluation routes.

Router learning uses paired plans from 264 development routes. Agent post-training and self-scheduling use 476 episodes partitioned by route into training and validation sets. Appendices~\ref{supp:data}--\ref{supp:timing} detail the splits, data collection, and training settings.

\paragraph{Implementation.}
We use a Qwen3-VL-2B backbone~\cite{qwen3vl} with a rank-64 task adapter. The Router has 0.84M parameters across two five-member ensembles and reads 64 BEV tokens plus context features. Routing updates every four simulation control ticks, and execution reports state every five ticks.

\subsection{Bench2Drive-220 performance}
\label{sec:main_results}
Tables~\ref{tab:main} and~\ref{tab:main_v004} compare models on Bench2Drive v0.0.3 and v0.0.4, respectively. \awa{} achieves state-of-the-art performance among PDM-Lite-trained models on both, obtaining 94.48 DS and 85.45\% SR on v0.0.3, and 92.33 DS and 82.73\% SR on v0.0.4. Table~\ref{tab:abilities} reports success rates across the five official abilities. \awa{} leads every ability among the listed methods, with mean ability SRs of 89.48\% on v0.0.3 and 84.79\% on v0.0.4. The following comparisons use v0.0.3 to examine pathway selection, agent learning, and execution feedback.

\subsection{Native pathways and learned selection}
\label{sec:ablations}
Table~\ref{tab:paradigms} compares fixed A, VLA, WAM, and the Router on all 220 routes with skills disabled to isolate computation choice.

The fixed pathways have complementary strengths (Fig.~\ref{fig:paradigms}a). A leads overtaking at 85.30 DS, VLA emergency braking at 93.00, and WAM merging and traffic signs at 85.83 and 87.17. WAM has the highest overall fixed-pathway DS of 87.91.

Learned routing improves overall driving. It reaches 89.70 DS and 73.64\% SR, gains of 1.79 DS and 4.09 percentage points over the best fixed scores. Its SR matches or exceeds the best fixed pathway in every ability group. It also exceeds the fixed-pathway average DS in four groups (Fig.~\ref{fig:paradigms}b), with comparable giving-way DS (84.80 versus 84.90).

\subsection{Agent learning and execution feedback}
Table~\ref{tab:ablation} ablates agent learning and execution feedback. The same 60 routes are used in all conditions and were selected to cover all 44 scenario types. Their full-model reference comes from the corresponding 220-route results and achieves 95.09 DS and 90.00\% SR.

\paragraph{Closed-loop agent learning.}
Replacing $G$ with its open-loop initialization reduces performance to 93.98 DS and 85.00\% SR, drops of 1.11 DS points and 5.00 percentage points of SR. The remaining learned components retain their full-model parameters. The result supports refining task decisions and two-view answers through closed-loop experience.

\paragraph{Event-triggered wake-up.}
Disabling event-triggered wake-up reduces performance to 92.57 DS and 85.00\% SR. Ordinary and mandatory event-triggered calls are removed, including rear-vehicle cues, and emergency yielding uses the perception-based rule. Periodic reviews still receive refreshed driving context and skill--target menus. The drops of 2.52 DS and 5.00 percentage points of SR support event-triggered calls when the scene or task changes.

\paragraph{Two-view queries.}
Disabling two-view queries reduces performance to 91.83 DS and 80.00\% SR, drops of 3.26 DS points and 10.00 percentage points of SR. This comparison replaces two-view queries and agent-based identity confirmation with perception-based emergency recognition and yielding. Other wake-up rules and the self-scheduler remain enabled. The result supports acquiring additional visual evidence in response to execution feedback.

A focused comparison on the five emergency-vehicle yielding routes raises mean DS from 70.0 to 100.0 when two-view queries are enabled. Figure~\ref{fig:police_case} illustrates the mechanism. A rear-vehicle cue prompts a two-view query, and $G$ identifies the approaching police vehicle and selects the rule skill to yield. It returns to native driving after the police vehicle passes.

These results support both directions of interaction between the model and execution. Agent post-training improves the decisions sent downward, while event-triggered requests and two-view queries improve when and with what evidence the agent is consulted. Learned routing addresses a complementary choice inside native planning by adapting which experts participate. Together, these mechanisms adapt the computation used for a trajectory and the task decision that governs execution.

\input{tab/runtime}
\input{tab/pathway_latency}

\subsection{Agent-call analysis}
Before learned self-scheduling, 72.4\% of 40,042 agent calls return an empty update or reaffirm the accepted skill--target choice (Table~\ref{tab:consultations}). This repetition motivates learning when to reconsider.

In offline replay of 89 route-disjoint validation episodes, the agent self-scheduler skips 2,629 of 7,313 calls (35.95\%) while retaining 959 of 1,002 calls labeled reconsider (95.71\%). Appendix~\ref{supp:timing} details the call labels and replay protocol. We next examine how routing and self-scheduling affect model computation.

\input{sec/5_efficiency}

%% file: tab/10_paradigms.tex
\begin{table*}[t]
\centering
\caption{\textbf{Native pathway selection on Bench2Drive v0.0.3.} All four configurations use the same model without skills. Each ability cell reports mean DS / success rate (\%). The last two columns report DS and SR over all 220 routes. Ability groups overlap. Higher is better. Bold highlights leading values.}
\label{tab:paradigms}
\small
\setlength{\tabcolsep}{3.2pt}
\begin{tabular}{lcccccrr}
\toprule
& \multicolumn{5}{c}{Ability-wise mean DS / SR (\%)} & \multicolumn{2}{c}{Overall} \\
\cmidrule(lr){2-6}\cmidrule(lr){7-8}
Pathway & Overtaking & Merging & \shortstack{Emergency\\brake} & \shortstack{Give\\way} & \shortstack{Traffic\\signs} & DS & SR (\%) \\
\midrule
Fixed A & 85.30 / \textbf{71.11} & 83.12 / 63.75 & 87.74 / 73.33 & \textbf{84.90} / \textbf{40.00} & 85.26 / 67.37 & 86.37 & 69.55 \\
Fixed VLA & 81.14 / 62.22 & 82.11 / 60.00 & \textbf{93.00} / \textbf{80.00} & \textbf{84.90} / \textbf{40.00} & 85.02 / 62.11 & 86.68 & 68.18 \\
Fixed WAM & 80.95 / 60.00 & 85.83 / 63.75 & 92.53 / \textbf{80.00} & \textbf{84.90} / \textbf{40.00} & 87.17 / 67.37 & 87.91 & 69.55 \\
\midrule
Learned Router & \textbf{86.54} / \textbf{71.11} & \textbf{87.93} / \textbf{70.00} & 92.30 / \textbf{80.00} & 84.80 / \textbf{40.00} & \textbf{88.02} / \textbf{69.47} & \textbf{89.70} & \textbf{73.64} \\
\bottomrule
\end{tabular}
\end{table*}

%% file: tab/3_ablations.tex
\begin{table}[!t]
\centering
\caption{\textbf{Ablations of agent learning and execution feedback.} All configurations use the same 60 selected Bench2Drive v0.0.3 routes.}
\label{tab:ablation}
\small
\begin{tabular*}{\linewidth}{@{\extracolsep{\fill}}lrr@{}}
\toprule
Configuration & DS $\uparrow$ & SR (\%) $\uparrow$ \\
\midrule
Without two-view queries & 91.83 & 80.00 \\
Without event-triggered wake-up & 92.57 & 85.00 \\
Open-loop $G$ & 93.98 & 85.00 \\
\midrule
\textbf{Full model} & \textbf{95.09} & \textbf{90.00} \\
\bottomrule
\end{tabular*}
\end{table}

%% file: fig/3_paradigms.tex
\begin{figure*}[t]
\centering
\includegraphics[width=\textwidth]{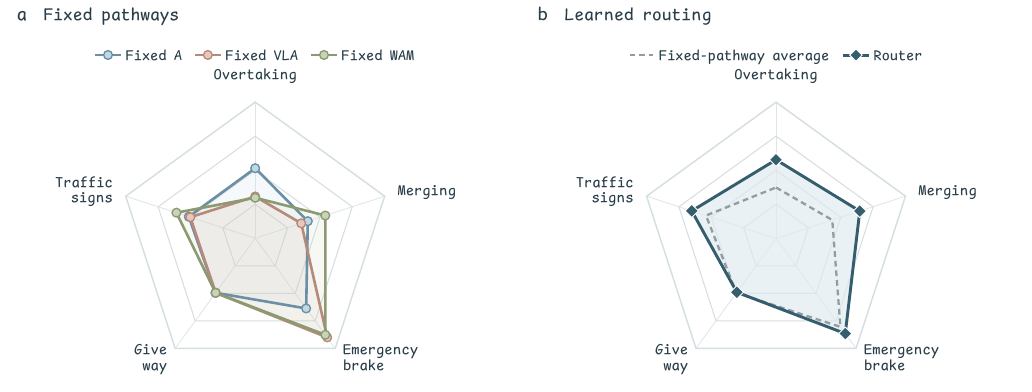}
\caption{\textbf{Routing combines complementary native capabilities.}
Mean driving score (DS) in each Bench2Drive v0.0.3 ability group, with skills disabled.
(a) Fixed A, VLA, and WAM. (b) Learned routing versus the mean of the three fixed pathways (dashed).
Both radars span 75--95 DS in 5-point intervals. Ability groups overlap. Table~\ref{tab:paradigms} gives the full values.}
\label{fig:paradigms}
\end{figure*}

%% file: tab/11_consultations.tex
\begin{table}[t]
\centering
\caption{\textbf{72.4\% of agent calls are empty updates or reaffirmations.} Agent-call outcomes across 476 episodes collected with sampled decoding before learned self-scheduling.}
\label{tab:consultations}
\small
\begin{tabular}{lrr}
\toprule
Call outcome & Calls & Share (\%) \\
\midrule
Empty update or reaffirmation & 29,006 & \textbf{72.4} \\
Other outcomes & 11,036 & 27.6 \\
\bottomrule
\end{tabular}
\end{table}

%% file: fig/4_police_case.tex
\begin{figure}[t]
\centering
\includegraphics[width=\columnwidth]{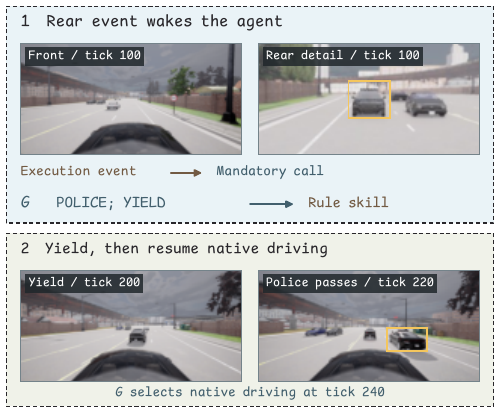}
\caption{\textbf{Execution feedback wakes the agent to yield to a police vehicle from behind.} A rear-vehicle cue prompts a two-view query bypassing the self-scheduler. $G$ activates the rule skill, then selects native driving after the police vehicle passes. All frames come from one recorded episode.}
\label{fig:police_case}
\end{figure}

%% file: tab/runtime.tex
\begin{table*}[t]
\centering
\caption{\textbf{Amortized model computation.} Values are estimated milliseconds per control tick using the same recorded Bench2Drive v0.0.3 invocation mix and RTX 5090 timings. The planning column includes one BEV encoder pass per sensor update, including updates during skill execution. Shared components comprise the detection and semantic heads, two-view queries, and traffic-signal queries. Totals use unrounded timings. Reductions are relative to fixed WAM with fixed-interval $G$.}
\label{tab:runtime}
\small
\setlength{\tabcolsep}{5pt}
\begin{tabular*}{\textwidth}{@{\extracolsep{\fill}}lrrrrr@{}}
\toprule
Configuration & \shortstack{Planning\\+ Router} & \shortstack{$G$\\+ self-scheduler} & Shared components & Total & Reduction (\%) \\
\midrule
Fixed WAM + fixed-interval $G$ & 137.23 & 58.55 & 13.47 & 209.25 & 0.00 \\
Router + fixed-interval $G$ & 80.02 & 58.55 & 13.47 & 152.03 & 27.34 \\
Fixed WAM + self-scheduling & 137.23 & 10.08 & 13.47 & 160.78 & 23.16 \\
\textbf{Full \awa{}} & 80.02 & 10.08 & 13.47 & \textbf{103.57} & \textbf{50.51} \\
\bottomrule
\end{tabular*}
\end{table*}

%% file: tab/pathway_latency.tex
\begin{table}[t]
\centering
\caption{\textbf{Complete native pathway latency and use.} Latency includes one BEV encoder pass and inference through the selected pathway, measured in milliseconds per plan on an RTX 5090 outside CARLA. Share is the percentage of native plans using each pathway on Bench2Drive v0.0.3.}
\label{tab:pathway_latency}
\small
\setlength{\tabcolsep}{5pt}
\begin{tabular*}{\columnwidth}{@{\extracolsep{\fill}}lrr@{}}
\toprule
Pathway & Share (\%) & Mean latency (ms) \\
\midrule
A & 54.84 & 38.87 \\
VLA & 10.52 & 157.96 \\
WAM & 34.65 & 176.35 \\
\bottomrule
\end{tabular*}
\end{table}

%% file: sec/5_efficiency.tex
\subsection{Computational efficiency}
\label{sec:runtime}
Routing and agent self-scheduling jointly reduce estimated model computation from 209.25 to 103.57\,ms per control tick, a 50.51\% reduction (Table~\ref{tab:runtime}). We profile warmed model calls on an RTX 5090 at batch size one with GPU synchronization. For the recorded Bench2Drive v0.0.3 workload, amortized computation sums each component's mean latency multiplied by its invocation count and divides by the number of control ticks. The estimate excludes input preprocessing, transport, and execution-layer CPU work. The Router reuses the BEV features from the current sensor update.

The four configurations use the same recorded workload to separate the computation savings from routing and self-scheduling. The baseline uses WAM for native planning and calls $G$ at every five-tick state report. We apply the recorded pathway choices, agent-call schedule, or both, with the same native planning frequency and the same counts of two-view and traffic-signal queries.

\paragraph{Router.}
A accounts for 54.84\% of native plans (Table~\ref{tab:pathway_latency}). The observed pathway mix averages 99.03\,ms per native plan, compared with 176.35\,ms for fixed WAM. Per control tick, fixed WAM costs 137.23\,ms and learned routing costs 80.02\,ms, including 1.76\,ms for the Router. The net saving is 57.22\,ms per control tick.

\paragraph{Agent self-scheduling.}
The agent self-scheduler keeps the current decision for 40.24\% of ordinary wake-up requests. Event-triggered requests and learned keep decisions together reduce task calls to $G$ by 82.84\% relative to fixed-interval calling. Including the self-scheduler's 0.04\,ms overhead per control tick, the combined cost of task calls and self-scheduling falls from 58.55 to 10.08\,ms, saving 48.47\,ms.

%% file: sec/6_conclusion.tex
\section{Conclusion}
\awa{} combines three native driving pathways with an agent that selects tools and learns when to reconsider. Dual-loop evolution improves these choices through closed-loop learning and failure-driven refinement of skills and execution feedback. The model achieves state-of-the-art performance among PDM-Lite-trained models on both Bench2Drive versions.

%% file: sec/llm_usage.tex
\section*{LLM Usage Statement}
Codex (GPT-6-Sol) and Claude Opus 5.5 assisted system implementation, analysis of execution records, code revision, and evaluation runs. GPT-6 assisted open-loop skill--target annotation and supervision-label review (Sec.~\ref{sec:rule}), as well as closed-loop failure attribution (Sec.~\ref{sec:outer}).

%% file: supp/1_model_and_data.tex
\section{Model and execution configuration}
\label{supp:model}
Agent, world, and action experts form three native pathways. A uses the action expert, VLA adds the agent expert, and WAM uses all three. A BEV encoder processes RGB and LiDAR. The Router selects expert activation from BEV tokens and compact context features. Native planning and task decisions share one vision-language backbone with role-specific LoRA adapters and separate key--value caches. System scheduling combines task selection by $G$ with call timing from the agent self-scheduler. Table~\ref{stab:model} summarizes the configuration.

\begin{table}[t]
\centering
\caption{\textbf{Model configuration.} World maps and trajectory points use the current ego coordinate frame.}
\label{stab:model}
\small
\begin{tabularx}{\columnwidth}{@{}lX@{}}
\toprule
Component & Configuration \\
\midrule
Sensors & Front RGB and LiDAR; additional rear RGB for two-view queries \\
Agent expert & Qwen3-VL-2B~\cite{qwen3vl}, 28 language layers, width 2048 \\
World expert & 28 Transformer layers, width 1024 \\
World output & Six $64\times64$ BEV maps, 13 semantic classes, 1\,m cells \\
Action expert & 28 Transformer layers, width 1024 \\
Action output & Six XY waypoints over 3\,s, a 20-point spatial route \\
Router & Learned pathway selection, 837,790 parameters \\
Task adapter & Rank 64, LoRA scaling $\alpha_{\rm LoRA}=128$, dropout 0.05, language attention and MLP projections \\
Planning history & Four frames at $(-0.75,-0.50,-0.25,0)$\,s \\
Agent-call history & Four frames at $(-1.5,-1.0,-0.5,0)$\,s, resized to $512\times256$ \\
Two-view query & Current front/rear views and motion context, at most 10 generated tokens \\
\bottomrule
\end{tabularx}
\end{table}

\paragraph{Native planning.}
The agent encodes camera history, speed, navigation targets, and a route command. Its layer-wise key--value states condition the action expert in VLA and the world expert in WAM. WAM conditions the action expert on both agent and world key--value states. A uses BEV features and ego/navigation inputs. All pathways share the action expert's Transformer layers, use pathway-specific input and output modules, and feed a common controller.

\paragraph{Shared parameters and adapters.}
Native reasoning and task decisions use separate LoRA adapters on the shared language backbone. VLA and WAM use the native-reasoning adapter, while task decisions and two-view queries use the task adapter. Routing and A use BEV inputs without native agent reasoning. Agent post-training updates the task adapter with all other model parameters fixed.

\paragraph{Task decisions.}
Each agent call combines images and perception-derived scene information with task state to form the driving context. The execution layer constructs the skill--target menu from observed objects, navigation context, and task state. At deployment, $G$ uses greedy decoding. The reply selects native driving or delegates a task for traffic negotiation, obstruction passing, navigation-goal adaptation, or traffic-rule compliance. The rule skill includes emergency-vehicle yielding. Task progress and the resulting scene inform subsequent agent calls.

\paragraph{Execution and agent calls.}
The simulation advances by 0.05\,s per control tick. Routing is refreshed every four ticks on planning requests. Execution reports driving state every five ticks. Changes in available skill--target choices, task and traffic events, and a review every 40 ticks produce wake-up requests from the execution layer. For ordinary requests, the agent self-scheduler selects reconsider (call $G$) or keep (skip the call). Mandatory requests call $G$ directly. The first wake-up request at least 120 ticks after the previous call invokes $G$ directly (the age-triggered refresh). Native planning and skill execution continue between agent calls. Section~\ref{sec:runtime} analyzes amortized model computation outside CARLA.

\paragraph{Execution feedback and two-view queries.}
\label{supp:view_input}
The execution layer rebuilds the driving context and skill--target menu from current sensor tracks, ego motion, navigation, and task state. Mandatory requests include execution interrupts and observed conflicts, meaning execution reports that the current task conflicts with measured scene conditions.

Active wake-up adds checks for persistent low progress without a traffic-control reason to wait, a new persistent blocker in the driving corridor, and newly available skill options that persist. These rules invoke $G$ with refreshed driving context and menu, bypassing the self-scheduler.

A rear-vehicle cue can additionally trigger a mandatory two-view query. The execution layer supplies current front and rear views, ego speed, and recent rear-track observations. Training and inference use the same rear-view crop. Section~\ref{supp:view_training} specifies the supervision.

\section{Data and partitioning}
\label{supp:data}
A, VLA, and WAM are trained jointly on driving logs in a mixed-pathway stream that also includes SimLingo question--answer examples~\cite{simlingo} and GPT-annotated skill--target choices for $G$. The open-loop Router is trained separately using pathway trajectory errors relative to PDM-Lite reference plans. Closed-loop CARLA data then refines routing, task selection, and self-scheduling. Each closed-loop record joins sensor features, skill--target menus, answers, task transitions, controls, and official infractions at a decision point.

Table~\ref{stab:data} summarizes the data and splits. All episodes from a route remain in one split, separate from the benchmark evaluation routes. Router configuration selection uses 223 training routes and 41 validation routes. The benchmark Router is retrained on all 264 development routes. The agent self-scheduler uses the same 120 training routes and 28 validation routes as $G$.

\begin{table}[t]
\centering
\caption{\textbf{Collected training and validation data.} Each row specifies its counting unit. Selected intervention points are defined in Sec.~\ref{supp:agent}.}
\label{stab:data}
\small
\begin{tabular}{lrr}
\toprule
Data & Train & Validation \\
\midrule
Router routes & 223 & 41 \\
Router replay contexts & 107,068 & 32,669 \\
Agent collection routes & 120 & 28 \\
Agent collection episodes & 387 & 89 \\
Logged agent calls & 32,729 & 7,313 \\
Selected intervention points & 308 & 73 \\
Reference driving examples & 554 & --- \\
\bottomrule
\end{tabular}
\end{table}

Task events and infractions are aligned to their occurrence frames. Agent post-training uses answer targets from the training partition and 554 reference driving examples for KL replay. Self-scheduler thresholds are calibrated on the training routes.

%% file: supp/2_router.tex
\section{Router learning with one-step offline RL}
\label{supp:router}
After initialization from open-loop trajectory errors against PDM-Lite plans, the Router learns an offline one-step decision policy using the contextual-bandit formulation from RL~\cite{levine2020offline,brandfonbrener2021offline,dudik2011doubly}. The context is a recorded observation, the action is a pathway choice, and the reward scores its three-second plan. We evaluate A, VLA, and WAM with paired sampling seeds, obtaining full-information supervision for all three choices. The deployed policy reads BEV tokens and context features.

\subsection{Paired pathway targets}
At each recorded context, we generate a three-second plan from each pathway with four paired seeds. The scorer applies driving criteria for collision, traffic lights, stop signs, lane departure, and progress, together with expert-trajectory deviation and nominal per-pathway latency measured in profiling. WAM includes world-guided plan refinement in both target construction and execution, and its latency includes that cost.

For context $i$, pathway $k$, and seed $m$, let $S_{i,k,m}$ be the change in route driving score from the recorded decision point to the end of the three-second plan replay. The scorer combines route completion, infraction penalties, and lane departure using the official driving-score criteria. Averaging the paired samples gives $\overline S_{i,k}$. We write the main paper's reward $r^R(x_i,k)$ as $r_{i,k}$ here. The reward is
\begin{equation}
 r_{i,k}=\overline S_{i,k}-\lambda_d\min(\overline d_{i,k},3)
               -\lambda_c\ell_k,
 \label{seq:router_reward}
\end{equation}
with score change in driving-score points, distance in meters, and nominal per-pathway latency in seconds. Training uses the weights in Table~\ref{stab:router_training}.

\subsection{Router architecture}
The policy attends to 64 BEV tokens, each projected from 1512 to 64 dimensions by a fixed PCA transform. A 25-dimensional context supplies the attention query through a 13-way situation indicator, 11 scalars, and a data-source indicator that is fixed at deployment. The scalars encode speed, one-second speed change, upcoming turn angle, junction status, lateral route deviation, red-light status, distance to turn onset, acceleration, front-object gap, gap availability, and following status. Situation categories distinguish junction maneuvers, stopping causes, obstacle avoidance, following, merging, vulnerable road users, and cruising. These features are computed from perception, ego motion, and navigation context.

A 64-dimensional query attends to projected token keys and values. The resulting summary and context feed two 96-unit hidden layers and a routing readout, with ReLU activations and dropout 0.2 in the hidden trunk. The two five-member ensembles of the Router contain 837,790 trainable parameters in total, excluding fixed PCA and normalization buffers.

The primary Router, the three-way stage that precedes the A--VLA refinement described below, factors its choice according to expert activation. A uses only the action expert, whereas VLA and WAM both activate the agent expert, and WAM additionally activates the world expert. A gate predicts the probability $g_i$ of activating agent reasoning, and a conditional head predicts the probability $h_i$ of VLA given that agent reasoning is active. The pathway probabilities are
\begin{equation}
 (p_{i,\mathrm A},p_{i,\mathrm{VLA}},p_{i,\mathrm{WAM}})
 = (1-g_i,\;g_i h_i,\;g_i(1-h_i)).
\end{equation}
This factorization separates whether to activate agent reasoning from whether to add world prediction. Both probabilities are predicted from the same context and trained jointly with the objectives below.
For training, rewards are centered within each context and clipped to the training-set 1st and 99th percentiles, giving $\widetilde r_{i,k}$. With $\operatorname{sg}$ denoting stopped gradients, the gate compares A with the VLA/WAM mixture defined by the current $h_i$:
\begin{align}
 v_i^A &= \widetilde r_{i,\mathrm A},\\
 v_i^D &= \operatorname{sg}(h_i)\widetilde r_{i,\mathrm{VLA}}
       +(1-\operatorname{sg}(h_i))\widetilde r_{i,\mathrm{WAM}},\\
 v_i^* &= \max(v_i^A,v_i^D).
\end{align}
The gate minimizes the expected reward lost by choosing A or this mixture instead of the better of the two:
\begin{equation}
 \mathcal L_{\rm gate}=\E_i[(1-g_i)(v_i^*-v_i^A)+g_i(v_i^*-v_i^D)].
\end{equation}
The conditional head learns which of VLA and WAM offers the higher reward, using binary cross-entropy with target $\mathbf 1[r_{i,\mathrm{VLA}}>r_{i,\mathrm{WAM}}]$. Examples are weighted by the absolute VLA--WAM reward gap and restricted to contexts where a deeper pathway improves on A. This focuses supervision on the value of world prediction when agent reasoning can improve driving reward, while giving near-tied choices little weight. Weights are normalized by their training-set mean. The two losses are optimized jointly.

The primary Router is a five-member ensemble and selects the pathway with the highest mean probability across its members. When it selects A, we ask whether enabling the agent expert alone justifies its extra latency once the reward also accounts for ego-trajectory comfort. A and VLA differ by precisely this expert. Switching from A to WAM would also enable world prediction, so the primary Router retains that decision.

On training contexts initially assigned to A, a five-member refinement ensemble learns the paired A-minus-VLA reward difference. Its target uses Eq.~\ref{seq:router_reward} with the refinement latency weight in Table~\ref{stab:router_training} and adds a comfort term. We standardize the comfort score on the training set, with higher values for smoother plans, and scale it by the training standard deviation of driving-score change before applying the comfort weight.

\subsection{Deployment}
Table~\ref{stab:router_training} summarizes the optimization settings. For an initial A choice, let $\mu$ and $\sigma$ denote the mean and standard deviation of the five predicted A-minus-VLA reward differences. The refinement retains A if $\mu+0.5\sigma>0$ and selects VLA otherwise. The Router selects among A, VLA, and WAM every four control ticks on native planning requests and retains that choice between updates.

\begin{table}[t]
\centering
\caption{\textbf{Router learning configuration.}}
\label{stab:router_training}
\small
\begin{tabularx}{\columnwidth}{@{}lX@{}}
\toprule
Setting & Value \\
\midrule
Optimizer & AdamW \\
Learning rate / weight decay & $10^{-3}$ / $10^{-4}$ \\
Batch size & 1024 \\
Routing objective & Expected regret + weighted BCE \\
Training passes per model & 8 \\
Latency weight $\lambda_c$ (primary / refinement) & 1.5 / 1.0 \\
Refinement comfort weight & 0.06 \\
Expert-deviation weight $\lambda_d$ & 2.0 \\
Selection training routes / contexts & 223 / 107,068 \\
Validation routes / contexts & 41 / 32,669 \\
Benchmark training routes / contexts & 264 / 139,737 \\
\bottomrule
\end{tabularx}
\end{table}

\subsection{Routing reward on validation routes}
\label{supp:regret}
We evaluate the Router trained on 223 routes using paired three-second plans from the 41 validation routes. We score the same plans twice with Eq.~\ref{seq:router_reward}, fixing the evaluation deviation weight at $\lambda_d=0.25$ and varying the evaluation latency weight $\lambda_c$ (Table~\ref{tab:router}). We average the paired reward differences over contexts and compute percentile intervals from 2,000 bootstrap samples of routes, retaining all contexts from each sampled route.

\input{tab/4_router}

%% file: tab/4_router.tex
\begin{table}[!t]
\centering
\caption{\textbf{Paired plan rewards on validation routes.} Entries show the mean reward difference between the Router and each indicated fixed pathway, with 95\% route-bootstrap intervals. Positive values favor the Router.}
\label{tab:router}
\small
\begin{tabular}{@{}lcc@{}}
\toprule
Latency weight & Router minus A & Router minus WAM \\
\midrule
$\lambda_c=0.1$ & $1.23$ $[0.49,2.50]$ & $0.11$ $[-0.02,0.23]$ \\
$\lambda_c=0.5$ & $1.20$ $[0.44,2.32]$ & $0.14$ $[0.01,0.27]$ \\
\bottomrule
\end{tabular}
\end{table}

%% file: supp/3_agent.tex
\section{Agent collection and post-training}
\label{supp:agent}
The collection records original agent recommendations, executed actions, randomized assignments, and task outcomes separately. These records supply task applicability targets and agent-call outcomes.

\subsection{Randomized intervention collection}
Randomized interventions broaden the recorded decisions beyond those the agent would choose itself. At episode initialization, a seeded draw assigns natural execution, forced call, or forced no-call with probabilities $0.5$, $0.25$, and $0.25$. An opportunity becomes eligible when an object first appears in the menu for a target skill. Each eligible opportunity is selected with probability $0.3$ until one is chosen, so an episode contributes at most one selected point.

The natural arm preserves the agent's sampled response. Collection uses full-vocabulary sampling at temperature $1.0$ with a budget of 128 new tokens, recording generated tokens and their probabilities. The forced-call arm repeats the selected target request until acceptance or four seconds. The forced-no-call arm suppresses the target skill family for 20 seconds. Other execution sources are recorded to identify compliance with the assigned intervention. Across the collection, 381 selected points receive 211 natural, 83 forced-call, and 87 forced-no-call assignments.

Outcome records align official progress and infractions, task completion, and the return to native driving with each agent call.

\subsection{Task applicability supervision}
Offline labeling rules use scenario annotations, the recorded skill--target menu, and task state to identify applicable skills. Obstruction-passing targets distinguish lane-blocking accident, construction, parked-vehicle, and door-opening scenarios. Navigation-goal adaptation targets identify an exit task, and emergency-yield targets identify the designated emergency vehicle. At deployment, the agent selects from the skill--target menu using images, perception-derived scene information, and task state.

Answer targets either delete an inapplicable first skill--target choice or keep an applicable recorded answer. Teacher forcing covers the full target. A second target operates on the first answer token. We use the task adapter from the preceding round of closed-loop training as the fixed reference. In the following losses, $x$ denotes the complete input to $G$, including the driving context and skill--target menu. Let $\pi_0$ denote the fixed reference policy, $\pi_0^{(1)}(v\mid x)$ its first-token distribution over vocabulary tokens $v$, and $b_x(v)$ an indicator for token prefixes associated with an inapplicable skill. The soft target is
\begin{equation}
 q_x(v)=\frac{\pi_0^{(1)}(v\mid x)\exp[-4b_x(v)]}
 {\sum_{v'}\pi_0^{(1)}(v'\mid x)\exp[-4b_x(v')]}.
\end{equation}
Minimizing $\KL(q_x\|\pi_G^{(1)})$ suppresses the unsupported prefixes while retaining a distribution over alternatives. Prefix groups handle alternative tokenizations of the same skill name.

We mix the closed-loop training records with a replay set of 554 reference driving examples. Replayed examples are trained solely through token-level KL to the fixed reference agent to retain prior responses across driving contexts. On closed-loop records, reference KL follows the original sampled answer prefixes. First-token examples apply it to the remaining answer positions. Deletion examples use only answer cross-entropy on the edited answer. Table~\ref{stab:agent_training} gives the training configuration.

For each record, the reference KL is
\begin{equation}
 K_i=\sum_{t\in T_i}\KL\bigl(
 \pi_0(\cdot\mid x_i,y_{i,<t})\,\|\,
 \pi_G(\cdot\mid x_i,y_{i,<t})\bigr),
\end{equation}
where $y_{i,<t}$ is the recorded answer prefix and $T_i$ contains the answer positions selected above. The combined objective is
\begin{equation}
\begin{split}
 \mathcal L_G={}&-\E_{\mathcal D_{\rm ans}}\log\pi_G(y_i^+\mid x_i)
 +\alpha\E_{\mathcal D_{\rm first}}\KL(q_i\|\pi_G^{(1)})\\
 &+\beta_{\rm cl}\E_{\mathcal D_{\rm cl}}K_i
 +\beta_{\rm replay}\E_{\mathcal D_{\rm replay}}K_i.
\end{split}
\label{eq:g_update}
\end{equation}
Here $y_i^+$ is the corrected or preserved answer, $q_i=q_{x_i}$ is the first-token target, and $\pi_G^{(1)}$ is the first-token distribution. The datasets $\mathcal D_{\rm ans}$ and $\mathcal D_{\rm first}$ supply answer and first-token targets. The answer dataset includes the two-view examples. The closed-loop records $\mathcal D_{\rm cl}$ and reference driving examples $\mathcal D_{\rm replay}$ supply reference KL $K_i$. Table~\ref{stab:agent_training} specifies $\alpha$, $\beta_{\rm cl}$, and $\beta_{\rm replay}$.

\begin{table}[t]
\centering
\caption{\textbf{Agent post-training recipe.} The mixed stream trains task decisions and two-view queries through the task adapter.}
\label{stab:agent_training}
\small
\begin{tabular}{lr}
\toprule
Setting & Value \\
\midrule
Optimizer & AdamW \\
Learning rate & $9\times10^{-6}$ \\
Weight decay & 0 \\
Effective batch size & 32 \\
Warm-up updates & 20 \\
Gradient norm limit & 1.0 \\
Answer weight / first-token weight $\alpha$ & 1 / 1 \\
KL weight $\beta_{\rm cl}$, closed-loop records & 0.02 \\
KL weight $\beta_{\rm replay}$, replayed examples & 0.10 \\
First-token suppression margin & 4.0 \\
Reference driving examples & 554 \\
\bottomrule
\end{tabular}
\end{table}

\subsection{Two-view queries}
\label{supp:view_training}
For the inputs in Sec.~\ref{supp:view_input}, scenario metadata and visual review determine a vehicle-type and yielding target. Non-yielding followers supply at least three negative examples per yielding example, including police vehicles outside a designated emergency scenario. Two-view examples enter the mixed training stream with answer cross-entropy only.

\section{Learning when to reconsider}
\label{supp:timing}
The agent-call dataset contains 40,042 agent calls from 476 episodes, with 32,729 training and 7,313 validation calls. Table~\ref{stab:consultation_outcomes} groups calls by updates to accepted skill--target choices and task state.
\begin{itemize}
\item \textbf{Empty update.} no choice update and unchanged task state.
\item \textbf{Reaffirmation.} existing choices are kept, possibly alongside rejected proposals.
\item \textbf{Choice change.} a skill--target choice is added, withdrawn, or revised.
\item \textbf{Rejected-only proposal.} a proposal is rejected without a kept or changed choice.
\item \textbf{Task transition.} task state changes without a change to the accepted choices.
\end{itemize}
If outcomes overlap, we assign the first matching category in the order choice change, task transition, reaffirmation, rejected-only proposal, and empty update. Choice changes and rejected-only proposals receive reconsider labels ($m_j=1$), and the remaining classes receive keep labels ($m_j=0$).

\begin{table}[t]
\centering
\caption{\textbf{Agent-call outcomes before learned self-scheduling.} Counts cover 40,042 agent calls.}
\label{stab:consultation_outcomes}
\small
\begin{tabular}{lrr}
\toprule
Outcome & Count & Share (\%) \\
\midrule
Empty update & 19,906 & 49.7 \\
Reaffirmation & 9,100 & 22.7 \\
Choice change & 2,944 & 7.4 \\
Rejected-only proposal & 2,654 & 6.6 \\
Task transition & 5,438 & 13.6 \\
\bottomrule
\end{tabular}
\end{table}

\paragraph{Training and calibration.}
The agent self-scheduler uses a lightweight classifier with features describing the wake-up request, elapsed time, scene dynamics, task state, and previous answer. It is trained with weighted binary cross-entropy, using positive weight $(1-\rho)/\rho$ for positive-label fraction $\rho$ in the fitting data. A threshold calibrated on training routes balances call reduction with retention of reconsider-labeled calls. The self-scheduler filters ordinary wake-up requests. Mandatory requests, the first agent call, and retries for missing or malformed answers invoke $G$ directly. Keep reuses the current task decision.

\paragraph{Held-out replay.}
Replay over 89 validation episodes on 28 routes includes 7,313 logged agent calls. The deployed self-scheduler skips 2,629 calls (35.95\%) and retains 959 of 1,002 reconsider-labeled calls (95.71\%). Of the 43 deferred reconsider-labeled calls, 42 have a subsequent agent call within the episode. The median interval to that call is five simulation control ticks. With the request-based refresh rule described in Appendix~\ref{supp:model}, the longest observed interval between calls was 160 ticks.

\section{Attribution and skill refinement}
\label{supp:outer}
Failure attribution uses trajectories, official events, selected pathways, agent decisions, task states, and controls. The attribution distinguishes errors in native action generation, pathway selection, agent decisions, and skill execution. It records events originating in the scenario separately. Skill-execution errors guide coding-agent changes to skills or execution feedback. The authors review these changes before rollouts with the refined skills supply new examples for the next inner-loop update.

%% file: supp/4_evaluation.tex
\section{Evaluation and additional analyses}
\label{supp:evaluation}
\subsection{Benchmark and controlled comparisons}
We use CARLA 0.9.15 and official seed 2000 for both Bench2Drive versions~\cite{carla,bench2drive,bench2drive004}. DS combines route completion with infraction penalties. SR is the percentage of routes with DS equal to 100. The five v0.0.3 ability groups contain 45 overtaking, 80 merging, 60 emergency-braking, 10 giving-way, and 95 traffic-sign routes. Routes can belong to multiple groups.

With skills disabled, we evaluate fixed A, VLA, WAM, and the Router on all 220 routes and report mean route DS and SR within each ability group. The three agent ablations share 60 routes covering all 44 scenario types. The full system scores 95.09 DS and 90.00\% SR on these routes. Replacing $G$ with its open-loop initialization removes closed-loop training for task decisions and two-view answers. The Router, native pathways, and self-scheduler keep their full-model parameters.

The wake-up ablation disables calls triggered by menu changes, task and traffic events, execution interrupts, observed conflicts, active wake-up rules, and rear-vehicle cues. Emergency yielding uses the perception-based rule. The first call, periodic reviews, and retries for missing or malformed answers remain enabled. Periodic reviews receive the updated driving context and skill--target menu. All learned parameters, the self-scheduler, and the age-triggered refresh retain their settings.

\subsection{Overlap in native pathway failures}
The native-pathway collection contains 430 scored runs on 139 distinct training routes. We have recorded runs for all three pathways on 81 routes. We define a failed run by official DS below 100. For each pathway, a route enters its failure set if any recorded run fails, pooling seeds and repeated runs. Across all 139 training routes, 61 had a recorded failure for at least one pathway. Of these, 32 (52.5\%) had a recorded failure for all three.